\documentclass{article}

\usepackage{microtype}
\usepackage{graphicx}
\usepackage{subfigure}
\usepackage{booktabs} 

\usepackage{hyperref}

\usepackage[accepted]{icml2024}

\usepackage{amsmath}
\usepackage{amssymb}
\usepackage{mathtools}
\usepackage{amsthm}

\usepackage[capitalize,noabbrev]{cleveref}

\theoremstyle{plain}

\theoremstyle{definition}

\theoremstyle{remark}

\usepackage[textsize=tiny]{todonotes}

\ifdefined\N                                                                
\renewcommand{\N}{\mathds{N}} 
\else \newcommand{\N}{\mathds{N}} \fi 
\newcommand{\R}{\mathds{R}} 
\ifdefined\C 
  \renewcommand{\C}{\mathds{C}} 
\else \newcommand{\C}{\mathds{C}} \fi

\newcommand{\argmin}{\operatorname{arg\,min}} 

\newcommand{\E}{\mathds{E}} 

\newcommand{\Xspace}{\mathcal{X}} 
\newcommand{\Yspace}{\mathcal{Y}} 
\newcommand{\Pxy}{\mathbb{P}_{xy}} 
\newcommand{\xv}{\mathbf{x}} 
\renewcommand{\xi}[1][i]{\xv^{(#1)}} 

\newcommand{\lamv}{\bm{\lambda}} 
\newcommand{\ind}{\mathcal{I}} 

\newcommand{\Hspace}{\mathcal{H}} 

\newcommand{\boldx}{\bm{x}}
\newcommand{\boldxobs}{\bm{x}^{(i)}}
\newcommand{\boldy}{\bm{y}}
\newcommand{\boldyobs}{\bm{y}^{(i)}}
\newcommand{\boldq}{\bm{q}}
\newcommand{\boldz}{\bm{z}}

\newcommand{\boldeta}{\bm{\eta}}

\newcommand{\Data}{\mathcal{D}}
\newcommand{\DataTrain}{\mathcal{D}_{\text{train}}}

\newcommand{\ntrain}{n_{\text{train}}}
\newcommand{\DataExplain}{\mathcal{D}_{\text{explain}}}

\renewcommand{\Pxy}{\mathbb{P}_{\boldx, \boldy}}

\usepackage{bm}
\usepackage{dsfont}
\usepackage{csquotes}

\icmltitlerunning{Position Paper: Bridging the Gap Between Machine Learning and Sensitivity Analysis}

\begin{document}

\twocolumn[
\icmltitle{Position Paper: \\
           Bridging the Gap Between Machine Learning and Sensitivity Analysis}



\icmlsetsymbol{equal}{*}

\begin{icmlauthorlist}
\icmlauthor{Christian A. Scholbeck}{lmu,mcml,arizona}
\icmlauthor{Julia Moosbauer}{lmu,mcml}
\icmlauthor{Giuseppe Casalicchio}{lmu,mcml}
\icmlauthor{Hoshin Gupta}{arizona}
\icmlauthor{Bernd Bischl}{lmu,mcml}
\icmlauthor{Christian Heumann}{lmu}
\end{icmlauthorlist}

\icmlaffiliation{lmu}{Department of Statistics, Ludwig-Maximilians-Universität in Munich, Munich, Germany}
\icmlaffiliation{mcml}{Munich Center for Machine Learning (MCML), Munich, Germany}
\icmlaffiliation{arizona}{Department of Hydrology and Atmospheric Sciences, The University of Arizona, Tucson AZ, USA}

\icmlcorrespondingauthor{Christian A. Scholbeck}{christian.scholbeck@lmu.de}

\icmlkeywords{Machine Learning, ICML}

\vskip 0.3in
]



\printAffiliationsAndNotice{}  

\begin{abstract}
    We argue that interpretations of machine learning (ML) models or the model-building process can be seen as a form of sensitivity analysis (SA), a general methodology used to explain complex systems in many fields such as environmental modeling, engineering, or economics.
    We address both researchers and practitioners, calling attention to the benefits of a unified SA-based view of explanations in ML and the necessity to fully credit related work.
    We bridge the gap between both fields by formally describing how (a) the ML process is a system suitable for SA, (b) how existing ML interpretation methods relate to this perspective, and (c) how other SA techniques could be applied to ML.
\end{abstract}

\section{Introduction}

Machine learning (ML) is concerned with learning models from data with applications as diverse as text \cite{zhang_textmining} and speech processing \cite{bhangale_speech}, robotics \cite{pierson_robotics}, medicine \cite{rajkomar_medicine}, climate research \cite{rolnick_climate}, or finance \cite{huang_finance}. Due to the increasing availability of data and computational resources, demand for ML has risen sharply in recent years, permeating all aspects of life. 
While the first publications in predictive modeling date back as far as the 1800s with Gauß and Legendre \cite{molnar_history, stigler_gauss_legendre}, the popularity of ML has surged in the twenty-first century, as it represents the current technological backbone for artificial intelligence. 
Increasing focus is put on interpretable models or the interpretation of black box models with model-agnostic techniques \cite{molnar_iml, rudin_iml_principles}, often referred to as interpretable ML (IML) or explainable artificial intelligence. Note that we utilize the term black box, although the internal workings of a model may be accessible but too complex for the human mind to comprehend. Furthermore, interpretations of the hyperparameter optimization (HPO) process have garnered attention in recent years \cite{hutter_fanova_automl}. In the context of this paper, we will refer to IML as any effort to gain an understanding of ML, including HPO.
\par
In a basic sense, sensitivity analysis (SA) \cite{saltelli_sa, razavi_future_sa, iooss_review_sa} is the study of how model output is influenced by model inputs. It is used as an assistance in many fields to explain input-output relationships of complex systems. Applications include environmental modeling \cite{gsa_hydrological, wagener_sa_earthsystem, shin_sa_hydrology, haghnegahdar_sa_hydrology, vrugt_sa_dependent, mai_sa_northamerica, nossent_sa_environmental}, biology \cite{sumner_sa_biology}, engineering \cite{guo_sa_engineering, ballesterripoll_sa_engineering, becker_sa_biomechanics}, nuclear safety \cite{saltelli_nuclear}, energy management \cite{tian_sa_energy}, economics \cite{harenberg_sa_economics, ratto_sa_dsge}, or financial risk management \cite{baur_gsa_finance}. In some jurisdictions such as the European Union, SA is officially required for policy assessment \cite{saltelli_sa_false}. With roots in design of experiments (DOE), SA started to materialize in the 1970s and 1980s with the availability of computational resources and the extension of DOE to design of computer experiments (DACE);
its large body of research is however spread across various disciplines, resulting in a lack of visibility \cite{razavi_future_sa}.
\par
\textbf{Why This Position Paper:} ML evolved largely independent of SA. As a result, the community did not fully credit related work and left potential research gaps unexplored. For instance, the high-dimensional model representation (HDMR) dates back to \citet{hoeffding_decomposition} and is the basis for variance-based SA, including Sobol indices \cite{sobol_index, homma_sa_sobol, rabitz_hdmr, saltelli_sa}. For the HDMR, decomposing the function into lower-dimensional terms is instrumental, an approach which was later redeveloped for ML and termed partial dependence (PD) \cite{friedman_pdp}. Furthermore, the HDMR is now better known as the functional analysis of variance (FANOVA) decomposition in ML \cite{hooker_fanova, hooker_generalizedfanova, molnar_iml} while many people are unaware of its roots in SA. Both the FANOVA \citep{hutter_fanova_automl} and the PD \citep{moosbauer_pd} have also been used to explain the HPO process. In algorithm configuration problems such as for HPO, evaluating paths of configurations by iteratively modifying parameters is known as ablation analysis
\cite{fawcett_ablation_hpo, biedenkamp_ablation_hpo_surrogates} but is strikingly similar to existing work in SA such as one-factor-at-a-time methods \cite{saltelli_sa}.
Recent advances in Shapley values \cite{strumbelj_shapley} have been made both in the SA \cite{owen_shapley_sobol} as well as the ML community \cite{lundberg_shap}. Several techniques have been developed to determine the importance of features in ML \cite{hooker_importance, casalicchio_featureimp, fisher_pfi, molnar_cpfi}, but numerous advances in other fields to compute variable importances \cite{wei_importance} are often overlooked. Grouping features for importance computations has only recently become relevant in ML \citep{au_grouped_importance}, although grouping system inputs has been an important topic in SA for decades \citep{sheikholeslami_grouping}. 

\textbf{Our Position:} We argue that IML can be seen as a form of SA applied to ML, which integrates recent advances in IML into a larger body of research on how to explain complex systems. We call attention to the benefits of a unified SA-based view of explanations in ML, to the necessity to fully credit related work, and to potential research gaps.
\par
To substantiate our claim, we bridge the gap between ML and SA by (a) formally describing the ML process as a system suitable for SA, (b) highlighting how existing methods relate to this perspective, and (c) discussing how SA methods\textemdash which are typically used in the domain sciences\textemdash can be applied to ML.


\section{Related Work}
In their survey of methods to explain black box models, \citet{guidotti_survey} mention SA as one approach besides other explanation methods, a distinction we aim to abandon in the context of this paper. \citet{razavi_future_sa} revisit the status of SA and conceive a vision for its future, including connections to ML regarding feature selection, model interpretations, and ML-powered SA; although they provide several clues about the connections between ML and SA, the paper does not formalize the interpretation process in ML and how it relates to SA in detail. \citet{scholbeck_framework} present a generalized framework of work stages for model-agnostic interpretation methods in ML, consisting of a sampling, intervention, prediction, and aggregation stage; although this methodology resembles SA (in the sense that an intervention in feature values is followed by a prediction with the trained model), the paper does not establish a formal connection. Several authors applied traditional SA methods to ML: \citet{fel_sobol_images} describe the importance of regions in image data with Sobol indices; \citet{kuhnt_gsa_ml} provide a short overview on variance-based SA for ML model interpretations; \citet{stein_gsa_xai} provide a survey of SA methods for model interpretations, conduct an analysis under different conditions, and apply the Morris method to compute sensitivity indices for a genomic prediction task; \citet{paleari_sa_morris} use the Morris method to rank the feature importance for a generic crop model;
\citet{tunkiel_sa_ml} use derivative-based SA to rank high-dimensional features for a directional drilling model; \citet{ojha_hpo_sa} evaluate the sensitivity of the model performance regarding hyperparameters using the Morris method and Sobol indices.
\par
Many authors have given considerable thought to the connection between ML and SA, either in the sense that SA can be applied to ML and vice versa or that methods in IML resemble some form of SA. We see this work here as connecting the dots between these efforts, establishing a formal link between ML and SA, directing attention to overlooked related work, and as a result, bringing both communities together.


\section{An Introduction to ML and SA}
\label{sec:background}

\subsection{Supervised Machine Learning}

In supervised learning, a model is learned from labeled data to predict based on new data from the same distribution with minimal error. To be precise, supervised learning requires a labeled data set $\Data = \left(\boldxobs, \boldyobs\right)_{i = 1, ..., n}$ of observations $\left(\boldxobs, \boldyobs\right)$ where $\boldxobs$ corresponds to the $p$-dimensional feature vector drawn from the feature space $\mathcal{X}$ and $\boldyobs$ to the $g$-dimensional target vector (also referred to as label) drawn from the target space $\mathcal{Y}$. We assume observations are drawn i.i.d. from an unknown distribution $\Pxy$ which is specific to the underlying learning problem:
$$
\left(\boldxobs, \boldyobs\right) \sim \Pxy
$$
We formalize the concept of training by introducing an inducer (or learner) $\ind$ as a function that maps the training subset $\DataTrain \subset \Data$ with $n_\text{train}$ observations and hyperparameter configuration $\lamv \in \Lambda$ to a model $\widehat{f}$ from a hypothesis space $\Hspace$:
\begin{align*}
    \ind: 
    \begin{cases}
        \Xspace \times \Yspace \times \Lambda &\to \Hspace  \\
        (\DataTrain, \lamv) &\mapsto \widehat{f}    
    \end{cases}
\end{align*}
Many learners use empirical risk minimization to train $\widehat{f}$:
\begin{align*}
    R_{\text{emp}}(\tilde{f}) &= \frac{1}{\ntrain} \sum_{i \,:\, (\boldx^{(i)}, \boldy^{(i)}) \,\in\, \DataTrain} L(\tilde{f}(\boldx^{(i)}), \boldy^{(i)}) \\
    \widehat{f} &= \argmin_{\tilde{f}} R_{\text{emp}}(\tilde{f})   
\end{align*}
$R_{\text{emp}}$ is only a proxy for the true generalization error (GE):
\begin{equation*}
    \mathrm{GE} =  \E_{(\boldx, \boldy) \sim \Pxy} \left[L(\widehat{f}(\boldx), \boldy) \right] 
\end{equation*}
$\widehat{f}$ is finally evaluated on an outer performance measure $\rho$ (which may coincide with $L$) on a test subset $\Data_{\text{test}} \subset \Data$:
\begin{align*}
\rho: 
    \begin{cases}
        \mathcal{X} \times \mathcal{Y} \times \mathcal{H} \to \mathbb{R} \\
        (\Data_{\text{test}}, \widehat{f}) \mapsto \widehat{\mathrm{GE}} 
    \end{cases}
\end{align*}
Experience has shown that resampling (and aggregating results) is a more efficient use of data; $\Data$ can be repeatedly split up into different train and test sets; $\DataTrain$ can be further split up in an inner loop to optimize hyperparameters \citep{bischl_hpo_review}.

\subsubsection{Hyperparameter Optimization}

In addition to controlling the behavior of the learner $\ind$, the entire learning procedure  is configurable by $\lamv$ which may, for example, control the hypothesis space (e.g., the number of layers of a neural network), the training process (e.g., a learning rate, regularization parameter, or resampling splits), or the data (e.g., a subsampling rate).
\par
We formalize the input-output relationship between $\lamv$ and the GE estimate as the function $c$:
\begin{eqnarray*}
    c: 
    \begin{cases}
        \Lambda &\to \R \\
        \lamv &\mapsto \widehat{GE}
    \end{cases}
\end{eqnarray*}
For most learning problems, this relationship is non-trivial, and it is of concern to select the hyperparameter configuration carefully. 
This has given rise to the HPO problem:
\begin{equation*}
    \lamv^\ast \in \argmin_{\lamv \in \Lambda} c(\lamv) \nonumber \\
\end{equation*}
In practice, one is typically interested in finding a configuration $\widehat \lamv$ with a performance close to the theoretical optimum $c(\widehat \lamv) \approx c(\lamv^\ast)$. Solving the optimization problem is challenging: usually, no analytical information about the objective function $c$ is available, evaluations of $c$ are expensive (a single evaluation of $c$ requires executing a training run), and the hyperparameter space might have a complex structure (hierarchical, mixed numeric-categorical). 
\par
Even experts find manual hyperparameter tuning through trial-and-error to be challenging and time-consuming, which creates a demand for HPO algorithms capable of efficiently discovering good solutions. Grid search (evaluating $c$ on an equidistant set of grid points in $\Lambda$) and random search (evaluating $c$ on a set of randomly sampled points in $\Lambda$) are the simplest approaches. More recently, Bayesian optimization \cite{jones98global} has become increasingly popular for hyperparameter tuning \cite{hutter11smac, snoek12practical}. 
\par
Without going into too much detail on the vast number of approaches in HPO \citep{bischl_hpo_review, feurer_hpo_review}, we simply denote a generic hyperparameter tuner by:
$$
\tau: 
    (\Data, \ind, \Lambda, L, \rho) \mapsto \widehat{\lamv}
$$
It maps the data $\Data$, inducer $\ind$, hyperparameter search space $\Lambda$, inner loss function $L$, and outer performance measure $\rho$ to the estimated best hyperparameter configuration $\widehat{\lamv}$.
\par
\textbf{Automated ML and Meta-Learning:}
The quality of an ML model is sensitive to many steps that need to be performed before actual training, including cleaning and preparation of a data set, feature selection or feature engineering, dimensionality reduction, selecting a suitable model class, and performing HPO. This process can be illustrated as a pipeline of work stages. The space of possible pipeline configurations can be seen as a (typically mixed-hierarchical) hyperparameter space, which HPO algorithms can optimize over, and which is the subject of automated ML.
\par
Also described as \enquote{learning to learn}, meta-learning is concerned with the study of how ML techniques perform on different tasks and using this knowledge to build models faster and with better performance \cite{vanschoren_metalearning}. In particular, one is interested in learning how task characteristics influence the behavior of learners.
Meta-learning can be considered an abstraction of HPO where the task characteristics represent hyperparameters.

\subsubsection{Interpretable Machine Learning}
\label{sec:iml_review}

\textbf{Model Interpretations:}
In recent years, explanations of the relationship between features and model predictions or features and the model performance have become an important part of ML \cite{molnar_iml}. There is no consensus regarding the definition or quantification of model interpretability, e.g., it can comprise sparsity of the model, the possibility of visual interpretations, decomposability into sub-models, and many other characteristics \cite{rudin_iml_principles}. Some model types can be interpreted based on model-specific characteristics (also referred to as intrinsically interpretable models), e.g., (generalized) linear models, generalized additive models, or decision trees. 
But many ML models, including random forests, gradient boosting, support vector machines, or neural networks, generally are black boxes. 
We can gain insights into the workings of such black boxes with model-agnostic techniques, which are applicable to any model type. 
\par
A general explanation process can be formalized as a function $\Gamma$ that maps the model $\widehat f$, explanation parameter configuration $\boldeta$, and a data set $\DataExplain$ to an explanation $\Xi$: 
\begin{align*}
    \Gamma: (\widehat{f}, \boldeta, \DataExplain) &\mapsto \Xi
\end{align*}
$\DataExplain$ can be used to query the model and may consist of training, test, or artificial feature values, as well as observed target values for loss-based methods \cite{fisher_pfi, casalicchio_featureimp, scholbeck_framework}.
Depending on the explanation method, $\boldeta$ may, for instance, control how to train a surrogate model or how to select or manipulate values in $\DataExplain$.
\par
$\Xi$ can be a scalar value: for instance, the permutation feature importance (PFI) \cite{fisher_pfi} and H-statistic \cite{friedman_rule_ensembles} indicate the importance of features or the interaction strength between features, respectively.
\par
$\Xi$ may also consist of a set of values indicating the effect of a subset of features on the predicted outcome, e.g., the individual conditional expectation (ICE) \cite{goldstein_ice}, partial dependence (PD) \cite{friedman_pdp}, accumulated local effects (ALE) \cite{apley_ale}, or Shapley values \cite{strumbelj_shapley, lundberg_shap}. Some methods adapt this methodology to evaluate the prediction loss \cite{casalicchio_featureimp}. Such functions, conditional on being lower-dimensional, also serve as visualization tools. 
\par
$\Xi$ can consist of data points for methods such as counterfactual explanations \cite{wachter_counterfactuals, dandl_counterfactuals}, which search for the smallest necessary changes in feature values to receive a targeted prediction. 
\par
Furthermore, some methods replace a complex non-interpretable model with a less complex interpretable one; here, $\Xi$ is a set of predictions returned by the surrogate. We differentiate between global surrogates, that train a surrogate on the entire feature space, or local ones, which do so for a single data point, e.g., local interpretable model-agnostic explanations (LIME) \cite{ribeiro_lime}.
\par
\textbf{Note that $\Gamma$ always provides information about how predictions of the trained model $\widehat{f}$ are influenced by the features}. A detailed illustration of this process for many model-agnostic techniques (which query the model with different feature values) such as the PFI, ICE, PD, ALE, LIME, or Shapley values is provided by \citet{scholbeck_framework}. The same holds, in general, for model-specific interpretations, which provide similar insight without querying the model. For instance, in linear regression models, beta coefficients are equivalent to certain sensitivity indices based on model queries \cite{saltelli_sa}.
\par 
\textbf{HPO Explanations:}
A distinct branch of IML is concerned with explanations of the HPO process (formalized by $c$). \citet{hutter_fanova_automl} compute a FANOVA of $c$ with a random forest surrogate model. \citet{moosbauer_pd} compute a PD of $c$ with uncertainty estimate enhancements. An ablation analysis \cite{fawcett_ablation_hpo, biedenkamp_ablation_hpo_surrogates} can be used to evaluate effects of iterative modifications of hyperparameters on the performance.
\citet{woznica_metalearning} explore the interpretation of meta models. 

\subsection{Sensitivity Analysis}

\subsubsection{Systems Modeling}

\begin{figure}[h]
    \centering
    \includegraphics[width = 0.35\textwidth]{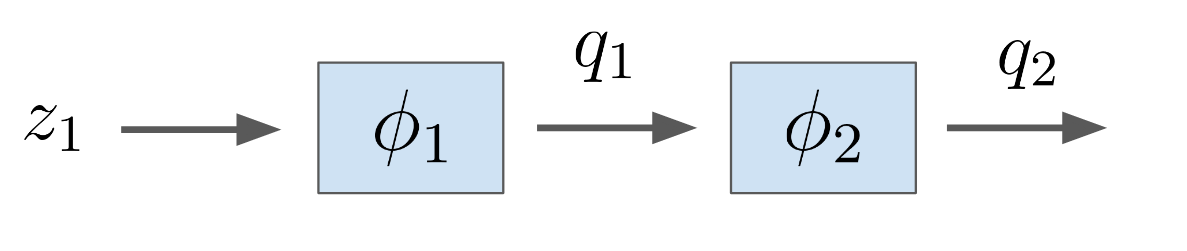}
    \caption{One option to visualize the example system in Eq. (\ref{eq:example_system}).}
    \label{fig:example_system}
\end{figure}

SA is concerned with modeling systems that consist of one or multiple interconnected models or functions \cite{razavi_future_sa} 
A model $\phi$ can either be determined manually or data-driven. The former is also referred to as law-driven, mechanistic, or process-based. A model receives an input vector $\boldz = (z_1, \dots, z_l) \in \mathbb{R}^l$ and returns an output vector $\boldq = (q_1, \dots, q_v) \in \mathbb{R}^v$:
\begin{align*}
    \phi(\boldz) &= \boldq
\end{align*}
As an illustration, consider a simple system that consists of two models where a model $\phi_2$ receives the scalar output of another model $\phi_1$ as an input:
\begin{align}
    \phi_1(z_1) &= q_1 \nonumber \\
    \phi_2(q_1) = q_2 \label{eq:example_system}
\end{align}
Systems are typically illustrated visually. As an example, consider the HBV-SASK hydrological system \citep{gupta_revisiting_esm}. One option is to visualize each model as a node that is connected to other nodes by streams of inputs and outputs (see Fig. \ref{fig:example_system}).
By building up a system of many smaller models instead of creating a single large model (quasi, creating a \enquote{network of models}), systems modeling allows for more sophisticated relationships between components while reducing complexity. As an example, consider an earth system model which could be composed of components modeling the atmosphere, ocean, land, and sea ice, and the exchange of energy and mass between these parts \cite{heavens_esm}. Whereas for a conventional singular climate model, one would require input data on carbon dioxide, earth system models can directly use anthropogenic emissions (caused by humans) and express the link between policies and climate change more explicitly \cite{kawamiya_esm}.

\subsubsection{SA of a System}

We analyze such a system for various purposes: assessing the similarity between a model and the underlying real phenomenon, determining the importance of input factors, identifying regions of the input space that contribute the most to output variability, evaluating the interdependence between input factors in their influence on the output, or identifying non-influential factors for the purpose of model simplification \cite{razavi_sa}.
\par
An input within the system (which we then manipulate for our analysis) can be any variable factor, e.g., model input values, model parameters, or constraints. We differentiate between local SA (LSA), which investigates a single location in the input space, and global SA (GSA), where input factors are varied simultaneously. In LSA, the inputs are deterministic, while in GSA,
$\boldz$ and $\boldq$ are considered random vectors with probability distributions \cite{borgonovo_sa}. 
LSA does not capture the aggregate behavior of a model, i.e., it can only provide insights into the influence of an input factor on the output for a single configuration (i.e., all remaining factors are kept constant). GSA has been developed with the motivation of explaining the model behavior across the entire input space. However, as opposed to local sensitivity, there is no unique definition of GSA methods \cite{razavi_sa}, which vary considerably in terms of their methodological approach \cite{iooss_review_sa, borgonovo_sa, razavi_future_sa}.
\par
The output of an SA method does not have to be identical to the output of a model. For instance, Monte-Carlo filtering produces random model outputs and identifies the ones that are located inside a region of interest (often referred to as the behavioral region) and the ones in the remaining regions (referred to as non-behavioral regions) \cite{saltelli_sa}. Monte-Carlo filtering is used in regional SA which aims to identify the most important input factors that lead to model outputs in the behavioral region. A \enquote{setting} determines how a factor's relevance or importance is defined and how it should be investigated, thereby justifying the use of certain methods for the task at hand. Common settings include factor prioritization and factor fixing, which aim to determine the most and least important input factors, respectively. The former is suited to rank the importance of input factors, while the latter is suited for screening input factors. 

\subsubsection{SA Methods}
SA methods can be categorized in multiple ways. This section provides an exemplary and non-exhaustive overview. The interested reader may be referred to the works of \citet{saltelli_sa}, \citet{borgonovo_sa}, \citet{pianosi_sa_review}, or \citet{razavi_future_sa} for further insights (or slightly different categorizations).
\par
\textbf{Finite-difference-based} methods aggregate finite differences (FDs) gathered at various points of the input space for a global representation of input influence. Input perturbations range from very small (numeric derivatives) to larger magnitudes. The Morris method, also referred to as the elementary effects (EE) method \cite{morris_method}, creates paths through the input space by traversing it one factor at a time, evaluating the model at each step of the path. Due to its low computational cost, the EE method is an important screening technique in SA to this date and has been modified numerous times \cite{saltelli_sa, campolongo_elementary}. One-factor-at-a-time methods are often criticized for leaving important areas of the feature space unexplored \cite{saltelli_oat}. A new generation of FD-based methods is referred to as derivative-based global sensitivity measures (DGSM) \cite{kucherenko_dgsm_sobol, sobol_dgsm, kucherenko_dgsm} which average derivatives at points obtained via random or quasi-random sampling. Variogram analysis of response surfaces \cite{razavi_variogram} is a framework to compute sensitivity indices based on the variance of FDs with equal distance across the input space. \citet{gupta_revisiting_esm} present a global sensitivity matrix consisting of derivatives w.r.t. each input for multiple time steps in dynamic systems. 
\par
\textbf{Distribution-based} methods aim to capture changes in the output distribution, often focusing on statistical moments such as the output variance (referred to as variance-based SA). In order to attribute the output variance to input effects of increasing order, the function to be evaluated is first additively decomposed into a high-dimensional model representation (HDMR) \cite{hoeffding_decomposition, sobol_index, saltelli_sa}. The fraction of explained variance by individual terms within the HDMR is referred to as the Sobol index \cite{sobol_index}. A link between DGSM and Sobol indices is demonstrated by \citet{kucherenko_dgsm_sobol}. Variance-based sensitivity indices can be estimated in various ways \cite{puy_sensitivity_estimators}. Recent efforts have focused on using Shapley values for variance-based SA, which can also be used for dependent inputs \cite{owen_shapley_sobol}. Contribution to the sample mean \cite{boladolavin_samplemean} and variance \cite{tarantola_samplevariance} plots visualize quantile-wise effects of inputs on the model output mean and variance. A common critique is that the simplification of the output distribution to a single metric such as the mean or variance entails an unjustifiable loss in information. Moment-independent techniques (which are also referred to as distribution-based methods by some authors) aim to capture changes in the entire output distribution and relate them to changes in input variables \cite{chun_moment_independent, borgonovo_moment_independent}.
\par
\textbf{Regression-based} methods are restricted to data-driven modeling. They either utilize some model-specific attribute, e.g., model coefficients, or evaluate the model fit w.r.t. changes in inputs, e.g., by excluding variables. For linear regression models, standardized correlation coefficients and partial correlation coefficients (which control for confounding variables) provide a natural sensitivity metric \cite{sudret_polynomial_chaos}. 
\par
\textbf{Emulators:} SA puts major focus on emulators or metamodels which approximate the model but are cheaper to evaluate. Such approximations are advantageous if the model is costly to evaluate and many model evaluations are needed such as in variance-based SA. Popular methods include Gaussian processes \cite{legratiet_pce_gp, marrel_gaussian_process, marrel_gaussian_process_sobol} and polynomial chaos expansion \cite{legratiet_pce_gp, sudret_polynomial_chaos}. Introducing a metamodel requires accounting for additional uncertainty in the metamodel itself and its estimation \cite{razavi_future_sa}.

\section{Bridging the Gap Between ML and SA}
\label{sec:bridginggap}

Evidently, there is a certain overlap between ML and SA: both fields are concerned with the explanation of input-output relationships, and many methods are used and developed concurrently, e.g., the FANOVA, model-specific explanations, evaluating changes in the model fit with varying features, or emulators. \textbf{Our position is that IML can be seen as a form of SA applied to ML.} Whereas SA is an explanation framework to analyze input-output relationships in virtually any complex system, we now analyze input-output relationships in a generalized ML system. 
We now formally describe this ML system and discuss how existing methods relate to this perspective and to each other.

\subsection{Viewing IML as a Form of SA Applied to ML}

Recall that in the SA (or systems modeling) sense, a model represents a function within the system. For the ML system we are modeling, this applies to the functions $\tau$, $c$, $\ind$, and $\Gamma$ which we defined in Section \ref{sec:background}. For instance, for $c$, we have that $\phi = c, \boldz = \lamv$, and $\boldq = \widehat{\text{GE}}$; for $\Gamma$, we have that $\phi = \Gamma, \boldz = (\widehat{f}, \boldeta, \DataExplain)$, and $\boldq = \Xi$.

\label{sec:ml_system}

\begin{figure}[h]
    \centering
    \includegraphics[width = \linewidth]{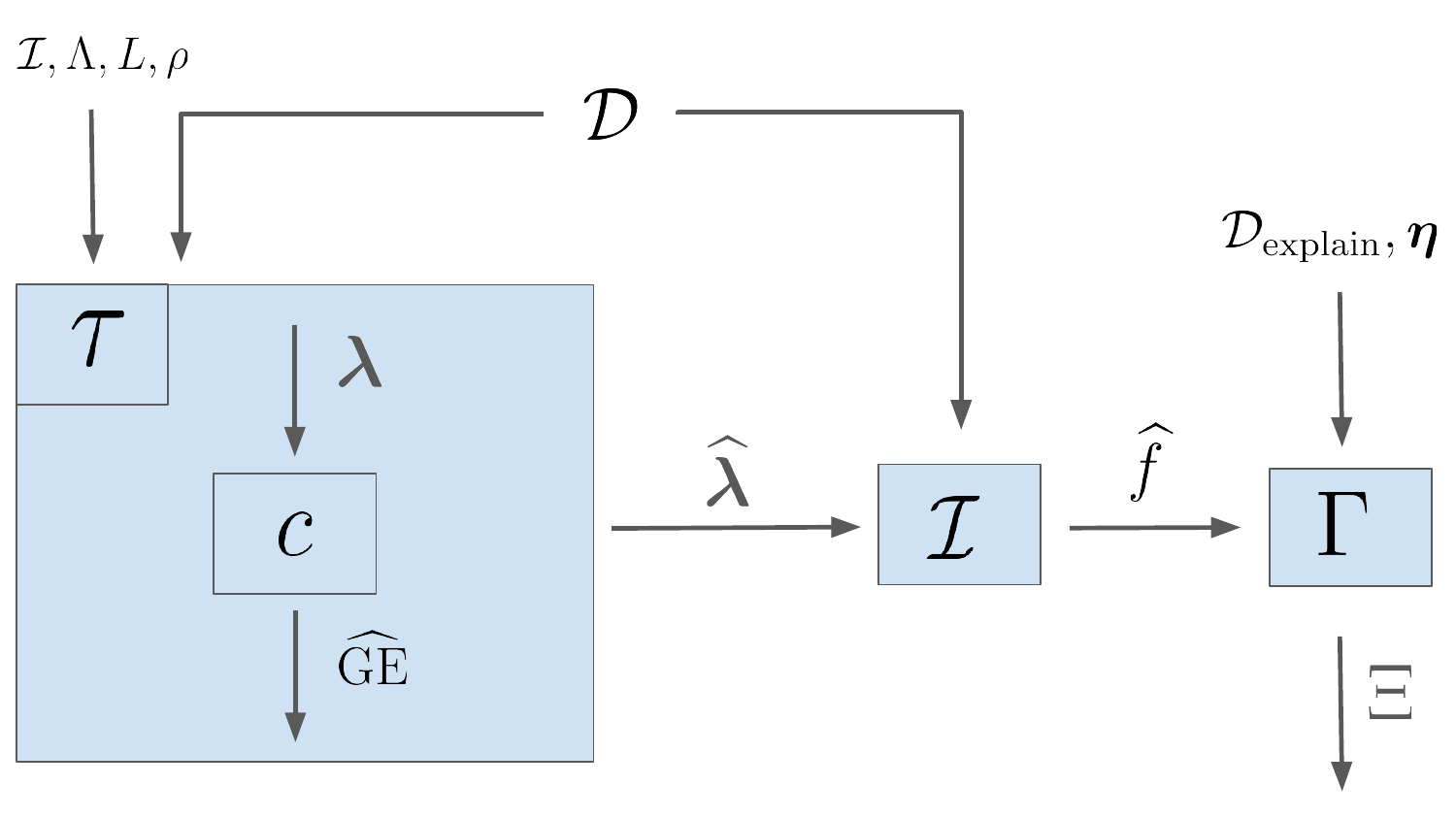}
    \caption{Formalizing the ML process as a system suitable for SA. The system consists of interconnected functions (indicated by boxes) of inputs and outputs.}
    \label{fig:ml_fsm}
\end{figure}

Fig. \ref{fig:ml_fsm} visualizes the ML system: functions correspond to boxes that receive inputs and produce outputs. This representation now enables us to view any explanation of the system under the common framework of SA.
We can see that there is a cascading or \enquote{trickle-down} effect: choosing different system inputs trickles down through multiple functions and results in various outputs throughout the system.
This might appear as an obvious fact, e.g., that choosing different hyperparameters results in a different model and consequently, in different model explanations, although the model is evaluated with the same data. \textbf{However, to formally describe sensitivities between different variables in ML, we first need to explicitly model their relationships.} 
In a certain sense, this is an effort to formalize a general theory of interpretations in ML.
\par
\textbf{IML Methods:} Recall that IML methods are formulated to operate on two levels: the model and the hyperparameter level. In the ML system, this corresponds to:
\begin{itemize}
    \item $\Gamma$: ICE, PD, FANOVA, ALE, PFI, counterfactual explanations, LIME, Shapley values 
    \item $c$: FANOVA \cite{hutter_fanova_automl}, PD \cite{moosbauer_pd}
\end{itemize}
Methods such as the PD and FANOVA have been used to explain the model and HPO process. After having presented a theory for SA within the ML system, this should not come as a surprise: every IML method simply computes sensitivities for input-output relationships (effects of features on the prediction, effects of features on the performance, or effects of hyperparameters on the performance) and could potentially be applied to other functions within the system. Furthermore, we can simply incorporate these novel IML methods into the SA toolbox and use them to compute sensitivities for entirely different systems, which, for instance, consist of physics-based models.
\par
\textbf{SA Methods:} One could potentially apply numerous other SA techniques to ML, which are typically used to explain models in fields associated with SA. This includes but is not limited to the Morris method, DGSM, the global sensitivity matrix, variograms / variogram analysis of response surfaces, or Sobol index estimators \cite{kuhnt_gsa_ml, stein_gsa_xai, paleari_sa_morris, tunkiel_sa_ml, ojha_hpo_sa, fel_sobol_images}.
\par
\textbf{Additional Input Parameters for SA:} In addition to using new methods to interpret the ML system, new interpretations can be created by considering different input parameters. Such suggestions are already often made in the context of specific methods, e.g., \citet{bansal_sensitivity_attributionmethods} analyze the sensitivity of model explanations for image classifiers regarding input parameters to the explanation method. Such novel types of SA for ML are now explicitly formalized in the context of this paper (with the explanation hyperparameter vector $\boldeta$ dictating how $\Gamma$ operates) and can easily be put into practice. For instance, we could evaluate the sensitivity of the PD curve when selecting different subsets of observations within $\DataExplain$ or the sensitivity of the surrogate prediction for a single instance $\boldx$ for variations in the width parameter for LIME.
\par
\textbf{Dependent Features and Adjustments:}
Traditional SA is often based on process-based models where the system can (in good faith) be queried with input space-filling designs. In ML, the model is typically trained with dependent features, resulting in many areas of the feature space where the model has not seen much or any data. 
Model predictions in such areas (typically referred to as extrapolations) do not reflect any underlying data-generating process. Furthermore, in variance-based SA, there is an additional error source for dependent features: the HDMR can be non-unique, and there are non-zero covariance terms in the variance decomposition \cite{chastaing_generalized_sobol}. 
\par
One might argue that in a model diagnostic sense, extrapolations do not pose a problem, as they represent accurate model predictions. Although there is some justification for this viewpoint, one might ask what the value behind a model and its explanations is if the model does not reflect any underlying real-world phenomenon. There are, however, potential solutions to this problem, at least for some methods: EEs or derivatives can, for instance, only be computed in high-density regions of the training set.

\subsection{Explanations on the Data Level}

Instead of evaluating function behavior, both SA and ML provide techniques that operate on the data directly. There are a few techniques that relate to traditional SA, e.g., SA on given-data \cite{plischke_given_data} (which aims to estimate sensitivity indices without querying a model) or traditional ML, e.g., the PD through stratification \cite{parr_pd_stratification} (which aims to do the same for the PD). Also referred to as green SA \cite{razavi_future_sa}, such sample-free approaches significantly reduce computational costs. Most data level techniques simply refer to some form of exploratory data analysis. For instance, scatter plots are used by the SA \cite{saltelli_sa} and ML \cite{hastie_elemstatlearn} communities for simple analyses.

\subsection{Model-Specific SA and Artificial Neural Networks}
\label{sec:ann_modelspecific_sa}

Model-specific interpretations typically represent some notion of sensitivity: for instance, tree splits tell us about changes in the average target value (the sensitivity of the target) when partitioning the data into subsets (w.r.t. inputs) for CART \citep{breiman_cart}; a beta coefficient for regression models informs us about the exact change in predicted outcome when adjusting feature values in a certain way. The nature of artificial neural networks (ANNs) allows for the design of powerful model-specific explanation techniques. Gradients can be more efficiently computed using symbolic derivatives which has resulted in numerous explanation methods \cite{ancona_gradients, neuralsens_package}.
ANNs have proven especially well-suited for unstructured data such as image, text, or speech data. Many ANN-specific explanation techniques have been designed for such data as well, e.g., saliency maps \cite{simonyan_saliency_maps}  visualize the sensitivity of the predicted target w.r.t. the color values of a pixel.
The term SA is often related to the analysis of ANNs \cite{yeung_sa_nn, zhang_sa_nn, neuralsens_package, mrzyglod_sa_nn}, much more so than in the context of analyzing other ML models. Ablation studies for ANNs \cite{meyes_ablation} analyze how the removal of certain components affects the model performance. As the range of model types and corresponding model-specific explanations is vast, they are not further discussed in this paper. However, we stress the importance of model-specific SA, if applicable.

\subsection{Further Contributions of SA}

Apart from providing methodological advances, SA can contribute to ML in a variety of other ways.
\par
\textbf{Best Practices:} The SA community has given considerable thought to what constitutes a high-quality analysis, which is typically termed sensitivity auditing \cite{saltelli_sensitivity_auditing, lopiano_sensitivity_auditing}. This includes answering questions on what the evaluated function or model is used for, what the assumptions are, reproducibility, or the viewpoint of stakeholders. Although such questions are also discussed in ML contexts, they have not been compiled and formalized in the same way as in SA. Furthermore, SA defines formal settings and definitions such as factor fixing and factor prioritization which are tied to setting-suitable metrics such as certain variants of Sobol indices. In ML, interpretation concepts are poorly defined and are often determined by the interpretation method itself. For instance, there is no definition of a ground-truth feature importance, and importance scores produced by different methods often cannot be compared. 
\par
\textbf{Application Workflows:}
In contrast to IML, the body of research in SA extensively discusses application workflows. For instance, it is common practice to screen for important input factors with computationally cheap methods first, e.g., using the Morris method or DGSM, before resorting to more accurate but computationally demanding techniques such as variance-based SA \cite{saltelli_sa}. As opposed to the research-focused, isolated development and evaluation of interpretation methods in ML, SA is much more embedded into industry applications and large-scale software systems. This has led to certain research directions that have been ignored in IML: for instance, \citet{sheikholeslami_grouping} explored strategies of grouping inputs to reduce computational costs; \citet{sheikholeslami_crashes} developed strategies to handle simulation crashes, e.g., due to numerical instabilities, without having to rerun the entire computer experiment.

\subsection{Contributions of ML to SA}

ML also significantly contributes to the advancement of SA.
Notable contributions include better metamodeling practices through novel ML models, dependence measures and kernel-based indices, and Shapley values \cite{razavi_future_sa}.
\par
In some fields such as hydrology, there still is a preference for theory-driven process-based modeling, which is often outperformed by modern ML models \cite{nearing_hydrology_ml}. 
Apart from providing better predictive performance in data-driven modeling, this indicates that the current understanding of certain real-world phenomena is insufficient and that there is a potential to ultimately create better theory-driven models. Major efforts are put into merging process-based and data-driven modeling (sometimes referred to as hybrid modeling) \cite{razavi_deep_learning, reichstein_process_based_modeling}. For instance, in one-way coupling, the output of a mechanistic model represents the input of an ML model; in modular coupling, system models are either created from process-based or data-driven modeling based on which approach better suits the sub-modeling task \cite{razavi_deep_learning}. 
\par
Furthermore, common constraints encountered in ML such as feature correlations have necessitated the development of interpretation methods able to handle these constraints, e.g., Shapley values. As noted by \citet{razavi_future_sa}, such developments are still immature in SA, which can profit from novelties in ML. There is a point to be made that every novel interpretation method designed in an ML context is also applicable to any other mathematical model (or system), e.g., physics-based ones. This is restricted to IML methods that only need access to predictions, which is the case for the majority of methods.
For instance, we imagine that ICEs, the PD, ALE, LIME, or Shapley values can provide tremendous value in interpreting various non-ML systems.

\section{Discussion}

In the following, major points for discussion shall be debated in a neutral light:
\par
\textbf{Why Should the ML Community Care About a Field With a Small Visibility?} This might be the reason that the ML community has given little attention to related work in SA. We would like to bring forward three arguments here: First, a sound scientific process should strive towards proper crediting related work and avoiding redundancies; second, viewing IML as a form of SA applied to ML clarifies how we think about existing methods and lets us establish a common framework and terminology to discuss and formulate methods (which, besides, can be argued irrespectively of SA as an independent discipline); third, recent developments demonstrate an untapped potential of applying existing SA methods to ML \cite{kuhnt_gsa_ml, stein_gsa_xai, paleari_sa_morris, tunkiel_sa_ml, ojha_hpo_sa}.
\par
\textbf{Is Every Method Included in This Framework?} We formulated a general system of ML suitable for SA that includes common model-agnostic methods \cite{scholbeck_framework} and novel methods to explain the HPO process, which have been directly derived from the SA literature, namely the FANOVA and the PD. Furthermore, we argued that this perspective includes model-specific interpretations such as tree splits, regression parameters, or ANN-specific methods. Even though this framework holds with great generality, we do not claim that every conceivable interpretation of the ML process is included. For the reasons we discussed in the previous paragraph, this does not, however, diminish the value of a unified SA-based perspective on IML.
\par
\textbf{What About Causality?} The question of whether a feature causes a change in the actual outcome and not just in the predicted outcome carries increasing significance for ML and cannot be answered based on the predictive model alone; additional assumptions are needed, e.g., in the form of a causal graph \cite{molnar_pitfalls}. Even if the model can perfectly predict the actual outcome, the causal effect can still run in both directions. Considerable effort has been put into the SA of causal inference (CI) models, e.g., to assess how robust associations are regarding unmeasured or uncontrolled confounding \cite{vanderweele_evalue, veitch_causal_sa, frauen_neural_sa}. Due to the complexity of CI and the limited scope of application (causality is only relevant in some applications whereas predictions are always relevant in ML), CI is not further discussed here. However, the relevance of SA for CI in the literature demonstrates that there is a potential overlap to be explored in the future.
\par
\textbf{What About Unsupervised Learning?}
We factored out unsupervised learning (UL), including clustering methods, from our analysis. This stems from the fact that UL has been mostly ignored by the recent trend in interpretability research. Notably, a few novel works explore SA-inspired methods for algorithm-agnostic cluster explanations, e.g., L2PC and G2PC \cite{ellis_g2pc} and feature attributions for clustering (FACT) \cite{scholbeck_clustering}. This further underpins our argument that SA is a unifying framework for interpretations in ML and may include UL. Future research in UL explanations will reveal their relationship with SA.

\section{Conclusion}

This paper aims to direct attention to the concurrent development of similar approaches to interpret mathematical models in multiple communities. Our position is that IML can be seen as a form of SA applied to ML, which integrates recent advances in IML into a larger body of research on how to explain complex systems. To further substantiate our claim, we formalize the ML process as a system suitable for SA and discuss how existing methods relate to this perspective. With this paper, we strive towards a better recognition of related work in the research community and the exploitation of potential research gaps.
\par
Some readers might agree, and some might disagree with our viewpoint. The nature of a position paper is to initiate a discussion, and if we achieve a change of direction within the ML community as outlined above, this paper will have fulfilled its purpose.

\bibliography{bibfile}
\bibliographystyle{icml2024}




\end{document}